\documentclass{article}

\usepackage[nonatbib,final]{neurips_2020}

\usepackage[utf8]{inputenc} 
\usepackage[T1]{fontenc}    
\usepackage{hyperref}       
\usepackage{url}            
\usepackage{booktabs}       
\usepackage{amsfonts}       
\usepackage{nicefrac}       
\usepackage{microtype}      

\usepackage{amsmath,amsfonts,bm}

\def\eqref#1{equation~\ref{#1}}

\def\1{\bm{1}}
\def\0{\bm{0}}

\def\vc{{\bm{\mathrm{c}}}}

\def\vx{{\bm{\mathrm{x}}}}

\def\vz{{\bm{\mathrm{z}}}}

\def\vtheta{{\bm{\theta}}}

\DeclareMathAlphabet{\mathsfit}{\encodingdefault}{\sfdefault}{m}{sl}
\SetMathAlphabet{\mathsfit}{bold}{\encodingdefault}{\sfdefault}{bx}{n}

\def\gX{{\mathcal{X}}}

\def\gZ{{\mathcal{Z}}}

\newcommand{\E}{\mathbb{E}}

\DeclareMathOperator*{\argmax}{arg\,max}

\usepackage{amsthm,xspace,xcolor,bbm}

\newtheorem{thm}{Theorem}

\newtheorem{prop}{Proposition}
\newtheorem{cor}{Corollary}

\newcommand{\Ind}[1]{{\ensuremath{\mathbbm{1}\!\left\{#1\right\}}}}

\newcommand{\distrib}{\ensuremath{P}\xspace}

\newcommand{\pdata}{{\ensuremath{\distrib_{r}}}\xspace}
\newcommand{\gen}{\ensuremath{g}\xspace}
\newcommand{\pgen}{{\ensuremath{\distrib_{\gen}}}\xspace}
\newcommand{\dis}{\ensuremath{d}\xspace}
\newcommand{\pdis}{{\ensuremath{\distrib_{\dis}}}\xspace}
\newcommand{\val}{\ensuremath{v}\xspace}

\newcommand{\valdis}{\ensuremath{a}\xspace}

\newcommand{\SL}[2]{\ensuremath{SL_{#1}\left(#2\right)}\xspace}

\usepackage{graphicx}
\usepackage{multirow}
\usepackage[utf8]{inputenc}
\usepackage[T1]{fontenc}

\title{Efficient Generation of Structured Objects with Constrained Adversarial Networks}
\author{%
    Luca Di Liello\thanks{Equal contributions.}\\
    University of Trento\\
    \And
    Pierfrancesco Ardino$^*$\\
    University of Trento\\
    \And
    Jacopo Gobbi$^*$\\
    University of Trento\\
    \AND
    Paolo Morettin\thanks{This work was partially carried out when PM was working at the University of Trento.}\\
    KU Leuven\\
    \\
    \And
    Stefano Teso\\
    University of Trento\\
    \texttt{firstname.lastname@unitn.it}\\
    \And
    Andrea Passerini\\
    University of Trento\\
}

\begin{document}
\maketitle

\begin{abstract}

    Generative Adversarial Networks (GANs) struggle to generate
    structured objects like molecules and game maps.
    The issue is that structured objects must satisfy hard requirements (e.g.,
    molecules must be chemically valid)
    that are difficult to acquire from examples alone.
    As a remedy, we propose Constrained Adversarial Networks (CANs), an
    extension of GANs in which the constraints are embedded into the model
    during training.
    This is achieved by penalizing the generator proportionally to the mass it
    allocates to invalid structures.
    In contrast to other generative models, CANs support efficient inference
    of valid structures (with high probability)
    and allows to turn on and
    off the learned constraints at inference time.
    CANs handle arbitrary logical constraints and leverage knowledge
    compilation techniques to efficiently evaluate the disagreement between the
    model and the constraints.
    Our setup is further extended to hybrid logical-neural constraints for
    capturing very complex constraints, like graph reachability.
    An extensive empirical analysis shows that CANs efficiently generate valid structures
    that are both high-quality and novel. 

\end{abstract}

\section{Introduction}

Many key applications require to generate objects that
satisfy hard structural constraints, like drug molecules, which must be
chemically valid, and game levels, which must be playable.
Despite their impressive
success~\cite{karras2018progressive,zhang2017stackgan,zhu2017unpaired},
Generative Adversarial Networks (GANs)~\cite{goodfellow2014generative} struggle in these applications.  The reason is
that data alone are often insufficient to capture the structural constraints (especially if noisy) and convey them to the model.
%

As a remedy, we derive Constrained Adversarial Networks (CANs), which
extend GANs to generating valid structures with high probabilty.
Given a set of arbitrary discrete constraints, CANs achieve this by
penalizing the generator for allocating mass to invalid objects during
training.  The penalty term is implemented using the semantic loss
(SL)~\cite{xu2018semantic}, which turns the discrete constraints into
a differentiable loss function implemented as an arithmetic circuit (i.e., a
polynomial).
The SL is probabilistically
sound, can be evaluated exactly, and supports end-to-end training.
Importantly, the polynomial -- which can be quite large,
depending on the complexity of the constraints -- can be thrown away
after training.
%
%
In addition, CANs handle complex constraints, like
reachability on graphs, by first embedding the candidate
configurations in a space in which the constraints can be encoded
compactly, and then applying the SL to the embeddings.

Since the constraints are embedded directly into the generator,
high-quality structures can be sampled efficiently (in time
practically independent of the complexity of the constraints) with a
simple forward pass on the generator, as in regular GANs.\footnote{With high probability.  Invalid structures, when generated, can be checked and rejected efficiently.  In this sense, CANs are related to learning efficient proposal distributions~\cite{azadi2018discriminator}.}  No costly
sampling or optimization steps are needed.  We additionally show how
to equip CANs with the ability to switch constraints on and off
dynamically during inference, at no run-time cost.

\textbf{Contributions.}  Summarizing, we contribute:  1)~CANs, an extension of
GANs in which the generator is encouraged at training time to generate valid
structures and support efficient sampling,
2)~native support for intractably complex constraints,
3)~conditional CANs, an effective solution for dynamically turning on and off the constraints at inference time,
4)~a thorough empirical study on real-world data showing that CANs generate structures that are likely valid and
coherent with the training data.

\section{Related Work}

Structured generative tasks have traditionally been tackled using probabilistic
graphical models~\cite{koller2009probabilistic} and
grammars~\cite{talton2012learning}, which lack
support for representation learning and efficient sampling under constraints.
Tractable probabilistic
circuits~\cite{poon2011sum,kisa2014probabilistic} are a recent
alternative that make use of ideas from knowledge
compilation~\cite{darwiche2002knowledge} to provide efficient
generation of valid structures.
%
These approaches generate valid objects by
  constructing a circuit (a polynomial) that encodes both the hard constraints and the
  probabilistic structure of the problem. Although inference is linear in the
  size of the circuit, the latter can grow very large if the constaints are complex enough.  In contrast, CANs model the probabilistic
  structure of the problem using a neural architecture, while relying
  on knowledge compilation for encoding the hard
  constraints during training.
  Moreover, the circuit can be discarded
  at inference time.
%
%
The time and space complexity of sampling for CANs is therefore
roughly independent from the complexity of the constraints in
practice.



Deep generative models developed for structured tasks are
special-purpose, in that they rely on ad-hoc architectures, tackle
specific applications, or have no support for efficient
sampling~\cite{guimaraes2017objective,de2018molgan,xue2019embedding,torrado2019bootstrapping}.
Some recent approaches have focused on incorporating a constraint
learning component in training deep generative models, using
reinforcement learning~\cite{de2018molgan} or inverse reinforcement
learning~\cite{hu2018deep} techniques. This direction is complementary
to ours and is useful when constraints are not known in advance or
cannot be easily formalized as functions of the generator output.
Indeed, our experiment on molecule generation shows the advantages of
enriching CANs with constraint learning to generate high quality and
diverse molecules.

Other general approaches for injecting knowledge into neural nets (like deep statistical-relational
models~\cite{lippi2009prediction,manhaeve2018deepproblog,marra2019neural}, tensor-based models~\cite{rocktaschel2017end,donadello2017logic}, and fuzzy logic-based models~\cite{marra2019lyrics}) are either not
generative or require the constraints to be available at inference time.

\section{Unconstrained GANs}

GANs~\cite{goodfellow2014generative} are composed of two neural nets:
a discriminator $\dis$ trained to recognize ``real'' objects
$\vx \in \gX$ sampled from the data distribution $\pdata$, and
a generator $\gen: \gZ \to \gX$ that maps random latent vectors $\vz \in \gZ$
to objects $g(\vx)$ that fool the discriminator.
Learning equates to solving the minimax game $\min_{\gen} \max_{\dis} \;
f_\text{GAN}(\gen, \dis)$ with value function:
\begin{equation}
    f_\text{GAN}(\gen, \dis) := \E_{\vx \sim \pdata}[\log \pdis(\vx)] + \E_{\vx \sim \pgen}[\log (1 - \pdis(\vx))]
    \label{eq:gan}
\end{equation}
Here $\pgen(\vx)$ and $\pdis(\vx) := \pdis(\text{real}\,|\,\vx)$ are the
distributions induced by the generator and discriminator, respectively.
%
%
New objects $\vx$ can be sampled by mapping random vectors $\vz$
using the generator, i.e., $\vx = \gen(\vz)$.
Under idealized assumptions, the learned generator matches the data
distribution:

\begin{thm}[\cite{goodfellow2014generative}]
    \label{thm:gan}

    If \gen and \dis are non-parametric and the leftmost
    expectation in Eq.~\ref{eq:gan} is approximated arbitrarily well by the
    data, the global equilibrium $(\gen^*, \dis^*)$ of
    Eq.~\ref{eq:gan} satisfies $\distrib_{\dis^*} \equiv \frac{1}{2}$ and
    $\distrib_{\gen^*} \equiv \pdata$.

\end{thm}

In practice, training GANs is notoriously
hard~\cite{salimans2016improved,mescheder2018training}.  The most
common failure mode is mode collapse, in which the generated objects
are clustered in a tiny region of the object space.  Remedies include
using alternative objective functions~\cite{goodfellow2014generative},
divergences~\cite{nowozin2016f,arjovsky2017wasserstein} and
regularizers~\cite{miyato2018spectral}.  In our experiments, we apply
some of these techniques to stabilize training.

In structured tasks, the objects of interest are usually discrete.  In the following, we focus on stochastic generators that output a \emph{categorical distribution} $\vtheta(\vz)$ over $\gX$ and objects are sampled from the latter.
In this case,
$
    \pgen(\vx) = \int_\gZ \pgen(\vx | \vz) p(\vz) d\vz = \int_\gZ \vtheta(\vz) p(\vz) d\vz = \E_{\vz}[\vtheta(\vz)]
$.

\section{Generating Structures with CANs}

Our goal is to learn a deep generative model that outputs structures $\vx$
consistent with validity constraints and an unobserved distribution \pdata.
We assume to be given:
%
%
i)~a feature map $\phi: \gX \to \{0,1\}^b$ that extracts $b$ binary features from $\vx$, and
ii)~a single validity constraint $\psi$ encoded as a Boolean formula on
$\phi(\vx)$.
If $\vx$ is binary, $\phi$ can be taken to be the identity; later we will discuss some alternatives.
Any discrete structured space can be encoded this way.

\subsection{Limitations of GANs}

Standard GANs struggle to output valid structures, for two main reasons.
First, the number
of examples necessary to capture any non-trivial constraint
$\psi$ can be intractably large.\footnote{The VC dimension of unrestricted discrete formulas is exponential in the number of variables~\cite{vapnik2015uniform}.}  This rules out learning the rules of chemical validity or, worse still, graph reachability from even moderately large data sets.  Second, in many cases of interest the examples are noisy and do violate
$\psi$, in which case the data lures
GANs into learning \emph{not} to satisfy the constraint:

\begin{cor}
    \label{thm:ideal_bad_gan}

    Under the assumptions of Theorem~\ref{thm:gan}, given a target distribution
    \pdata, a constraint $\psi$ consistent with it, and a dataset of examples
    $\vx$ sampled i.i.d. from a corrupted distribution $\pdata' \ne
    \pdata$ inconsistent with $\psi$, GANs associate non-zero mass to
    infeasible objects.

\end{cor}

This follows easily from Theorem~\ref{thm:gan}, as the optimal generator satisfies $\pgen \equiv
\pdata'$, which is inconsistent with $\psi$.
Since Theorem~\ref{thm:gan} captures the \emph{intent} of GAN training, this corollary shows that GANs are \emph{by
design} incapable of handling invalid examples.

\subsection{Constrained Adversarial Networks}

Constrained Adversarial Networks (CANs) avoid these issues by taking both the data and the target structural constraint $\psi$ as inputs.  The value function is designed so that the generator maximizes the probability of generating valid structures.  In order to derive CANs it is convenient to start from the following alternative GAN value function~\cite{goodfellow2014generative}:
$
    f_\text{ALT}(\gen, \dis) := \E_{\vx \sim \pdata}[\log \pdis(\vx)] - \E_{\vx \sim \pgen}[\log \pdis(\vx)]
    \label{eq:altgan}
$.

Let $(\gen, \dis)$ be a GAN and $\val(\vx) = \Ind{\phi(\vx) \models \psi}$ be a fixed discriminator
that distinguishes between valid and invalid structures, where
$\models$ indicates logical entailment.  Ideally, we wish the generator to \emph{never} output invalid structures.  This can be achieved by using an aggregate
discriminator $\valdis(\vx)$ that only accepts configurations that are both valid and high-quality w.r.t. $\dis$.  Let $A$ be the indicator that $\valdis$ classifies $\vx$ as real, and similarly for $D$ and $V$.  By definition:
\begin{equation}
    \distrib_\valdis(\vx)
    = \distrib(A \,|\, \vx)
    = \distrib(D \,|\, V, \vx) \distrib(V \,|\, \vx)
    = \pdis(\vx) \Ind{\phi(\vx) \models \psi}
    \label{eq:valdis}
\end{equation}
Plugging the aggregate discriminator into the alternative value function gives:
\begin{align}
    & \argmax_\valdis \; f_\text{ALT}(\gen, \valdis)
    \\
    & = \argmax_\dis \; \E_\pdata[\log \pdis(\vx) + \log \Ind{\phi(\vx) \models \psi}] - \E_\pgen[\log \pdis(\vx) + \log \Ind{\phi(\vx) \models \psi}]
    \\
    & = \argmax_\dis \; \E_\pdata[\log \pdis(\vx)] - \E_\pgen[\log \pdis(\vx)] - \E_\pgen[\log \Ind{\phi(\vx) \models \psi}]
    \\
    & = \argmax_\dis \; f_\text{ALT}(\gen, \dis) - \E_\pgen[\log \Ind{\phi(\vx) \models \psi}]
    \label{eq:altcan}
\end{align}
The second step holds because $\E_\pdata[\log \Ind{\phi(\vx) \models \psi}]$
does not depend on \dis.  If \gen allocates non-zero mass to \emph{any}
measurable subset of invalid structures, the second term becomes $+\infty$.
This is consistent with our goal but problematic for learning.  A better alternative is to optimize the lower bound:
\begin{equation}
    \SL{\psi}{\gen}
    :=
    -\log \pgen(\psi)
    =
    - \log \E_\pgen[\Ind{\phi(\vx) \models \psi}]
    \le
    - \E_\pgen[\log \Ind{\phi(\vx) \models \psi}]
    \label{eq:lowerbound}
\end{equation}
This term is the \emph{semantic loss} (SL) proposed in~\cite{xu2018semantic}
to inject knowledge into neural networks.  The SL is much
smoother than the original and it only evaluates to $+\infty$ if \pgen
allocates \emph{all} the mass to infeasible configurations.  This immediately
leads to the CAN value function:
\begin{equation}
    \textstyle
    f_\text{CAN}(\gen, \dis) := f_\text{ALT}(\gen, \dis) + \lambda \SL{\psi}{\gen}
    \label{eq:can}
\end{equation}
where $\lambda > 0$ is a
hyper-parameter controlling the importance of the constraint.  This formulation is related to integral probability metric-based GANs, cf.~\cite{li2017mmd}.  The SL can be viewed as the negative log-likelihood of $\psi$, and hence it rewards the generator
proportionally to the mass it allocates to valid structures.  The expectation in Eq.~\ref{eq:lowerbound} can be rewritten as:
\begin{equation}
    \E_{\vx \sim \pgen}[\Ind{\phi(\vx) \models \psi}]
    = \sum_{\vx : \phi(\vx) \models \psi} \pgen(\vx)
    = \E_{\vz}\left[ \sum_{\vx : \phi(\vx) \models \psi} \: \prod_{i \,:\, x_i = 1} \theta_i(\vz) \!\! \prod_{i \,:\, x_i = 0} (1 - \theta_i(\vz)) \right]
    \label{eq:wmc}
\end{equation}
Hence, the SL is the negative logarithm of a polynomial in $\vtheta$ and
it is fully differentiable.\footnote{As long as $P_\gen(\vx) > 0$, which is always the case in practice.}  In practice, below we apply the semantic loss term directly to $f_\text{GAN}$, i.e., $f_\text{CAN}(\gen,\dis) := f_\text{GAN}(\gen,\dis) + \lambda \SL{\psi}{\gen}$.

If the SL is given large enough weight $\lambda$ then it gets closer to the ideal ``hard'' discriminator, and therefore more strongly encourages the CAN to
generate valid structures.  Under the preconditions of
Theorem~\ref{thm:gan}, it is clear that for $\lambda \to \infty$ CANs
generate valid structures only:

\begin{prop}
    \label{thm:ideal_can}

    Under the assumptions of Corollary~\ref{thm:ideal_bad_gan}, CANs associate
    zero mass to infeasible objects, irrespective of the discrepancy between
    \pdata\ and $\pdata'$.

\end{prop}

Indeed, any global equilibrium $(\gen^*, \dis^*)$ of $\min_\gen
\max_\dis f_\text{CAN}(\gen, \dis)$ minimizes the second term: the minimum is attained by $\log
\distrib_{\gen^*}(\psi) = 0$, which entails $\distrib_{\gen^*}(\lnot \psi) = 0$.
Of course, as with standard GANs, the prerequisites are often violated in
practice.  Regardless, Proposition~\ref{thm:ideal_can} works as a sanity check,
and shows that, in contrast to GANs, CANs are appropriate for structured
generative tasks.

\begin{figure}[t]
    \centering
    \includegraphics[width=0.9\linewidth]{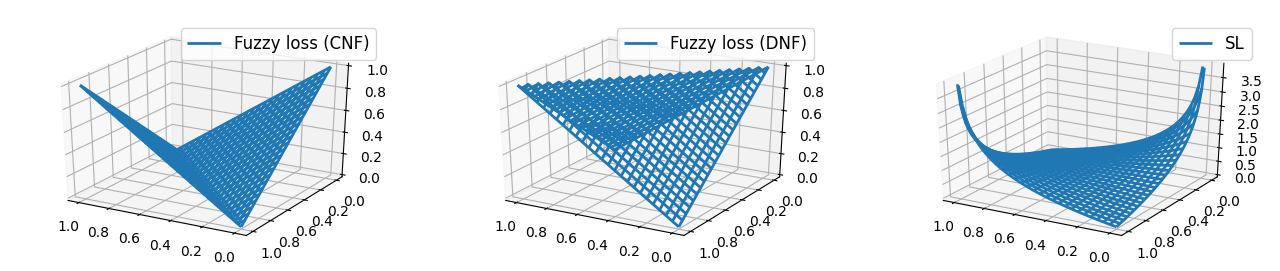}
    \caption{Left: fuzzy logic encoding (using the Łukasiewicz T-norm) of $x \oplus y$ in CNF format as a function of $P(x=1)$ and $P(y=1)$.
    Middle: encoding of DNF XOR.
    Right: SL of either encoding.}
    \label{fig:sl_vs_fl}
\end{figure}

A possible alternative to the SL is to introduce a differentiable knowledge-based
loss into the value function by relaxing the constraint $\psi$ using
fuzzy logic, as done in a number of recent works on discriminative deep
learning~\cite{donadello2017logic,marra2019lyrics}. Apart from lacking
a formal derivation in terms of expected probability of satisfying
constraints, the issue is that fuzzy logic is not semantically sound,
meaning that equivalent encodings of the same constraint may give
different loss functions~\cite{giannini2018convex}.
Figure~\ref{fig:sl_vs_fl} illustrates this on an XOR
constraint: the ``fuzzy loss'' and its gradient change radically
depending on whether the XOR is encoded as CNF (left) or DNF (middle),
while the SL is unaffected (right).

\paragraph{Evaluating the Semantic Loss}
%
%
The sum in Eq.~\ref{eq:wmc} is the unnormalized probability of
generating a valid configuration.  Evaluating it
requires to sum over \emph{all} solutions
of $\psi$, weighted according to their probability with respect to
$\vtheta$.  This task is denoted Weighted Model Counting
(WMC)~\cite{chavira2008probabilistic}.  Na\"ively implementing WMC is
infeasible in most cases, as it involves summing over exponentially
many configurations.  Knowledge
compilation (KC)~\cite{darwiche2002knowledge} is a well known approach in
automated reasoning and solving WMC through KC is the state-of-the-art
for answering probabilistic queries in discrete graphical
models~\cite{chavira2008probabilistic,fierens2015inference,van2011lifted}.
Roughly speaking, KC leverages distributivity to rewrite the polynomial in Eq.~\ref{eq:wmc} as compactly as possible, often offering a tremendous speed-up during evaluation.
This is achieved by identifying shared sub-components and compactly representing the factorized polynomial using a DAG.  Target representations for the DAG (OBDDs, DNNFs, \emph{etc.}~\cite{darwiche2002knowledge}) differ in succinctness and enable different polytime (in the size of the DAG) operations.  As done in~\cite{xu2018semantic}, we compile the SL polynomial into a Sentential Decision Diagram (SDD)~\cite{darwiche2011sdd} that enables efficient WMC and therefore exact evaluation of the SL and of its gradient.
%
%
KC is key in making evaluation of the Semantic Loss and of its gradient practical, at the cost of an offline compilation step -- which is however performed only once before training.

The main downside of KC is that, depending on the complexity of $\psi$,
the compiled circuit may be large.
This is less of an issue during training, which is often performed on
powerful machines, but it can be problematic for inference, especially
on embedded devices.  A major advantage of CANs is that the circuit is not
required for inference (as the latter consists of a simple forward pass over the
generator), and can thus be thrown away after training.  This means
that CANs incur no space penalty during inference compared to
GANs.

\paragraph{The embedding function $\phi$}  The embedding function $\phi(\vx)$ extracts Boolean variables to which the SL is then applied.  In many cases, as in our molecule experiment, $\phi$ is simply the identity map.  However, when fed a particularly complex constraint $\psi$, KC my output an SDD too large even for the training stage.  In this case, we use $\phi$ to map $\vx$ to an application-specific embedding space where $\psi$ (and hence the SL polynomial) is expressible in compact form.
We successfully employed this technique to synthesize Mario levels where the goal tile is reachable from the starting tile;  all details are provided below.  The same technique can be exploited for dealing with other complex logical formulas beyond the reach of state-of-the-art knowledge compilation.

\subsection{Conditional CANs}
\label{sec:infocan}

So far we described how to use the SL for enforcing structural
constraints on the generator's output. Since the SL can be applied to
any distribution over binary variables, it can also be used to enforce
conditional constraints that can be turned on and off at inference
time. Specifically, we notice that the constraint can involve also
latent variables, and we show how this can be leveraged for different
purposes.
Similarly to InfoGANs~\cite{chen2016infogan}, the generator's
input is augmented with an additional binary vector $\vc$. Instead of
maximizing (an approximation of) the mutual information between $\vc$
and the generator's output, the SL is used to logically bind the input
codes to semantic features or constraint of interest.
Let $\psi_1, \ldots, \psi_k$ be $k$ constraints of interest.  In order
to make them switchable, we extend the latent vector $\vz$ with $k$
fresh variables $\vc = (c_1, \ldots, c_k) \in \{0,1\}^k$ and train the
CAN using the constraint:
\[
    \textstyle
    \psi = \bigwedge_{i=1}^k (c_i \leftrightarrow \psi_i)
    \label{eq:onoff}
\]
where the prior $P(\vc)$ used during training is estimated from
data.

Using a conditional SL term during training results in a model that
can be conditioned to generate object with desired, arbitrarily
complex properties $\psi_i$ at inference time. Additionally, this
feature shows a beneficial effect in mitigating mode collapse
during training, as reported in Section~\ref{sec:molecules}.

\section{Experiments}

Our experimental evaluation aims at answering the following questions:
\begin{itemize}
\item[\textbf{Q1}] Can CANs with tractable constraints achieve better
  results than GANs?
\item[\textbf{Q2}] Can CANs with intractable constraints achieve better
  results than GANs?
\item[\textbf{Q3}] Can constraints be combined with rewards to achieve better results than using rewards only?
\end{itemize}
We implemented CANs\footnote{The code is freely available at https://github.com/unitn-sml/CAN} using Tensorflow and used PySDD\footnote{URL:
  \url{pypi.org/project/PySDD/}} to perform knowledge compilation.  We
tested CANs using different generator architectures on three real-world
structured generative tasks.\footnote{Details can be found in
  the Supplementary material.}
In all cases, we evaluated the objects generated by CANs and those of the
baselines using three metrics (adopted from~\cite{samanta2018designing}):
\textbf{validity} is the proportion of sampled objects that are valid;
\textbf{novelty} is the proportion of valid sampled objects that are not present in the training data; and
\textbf{uniqueness} is the proportion of valid unique (non-repeated) sampled objects.

\subsection{Super Mario Bros level generation}
In this experiment we show how CANs can help in the challenging
task of learning to generate videogame levels from user-authored
content. While procedural approaches to videogame level generation
have successfully been used for decades, the application of machine
learning techniques in the creation of (functional) content is a
relatively new area of research~\cite{summerville2018procedural}.
On the one hand, modern video game levels are characterized by
aesthetical features that cannot be formally encoded and thus are
difficult to implement in a procedure, which motivates the use of ML
techniques for the task. On the other hand, the levels have often to
satisfy a set of functional (hard) constraints that are easy to
guarantee when the generator is hand-coded but pose challenges for
current machine learning models.

Architectures for Super Mario Bros level generation include
LSTMs~\cite{summerville2016super}, probabilistic graphical
models~\cite{guzdial2016game}, and multi-dimensional
MCMC~\cite{snodgrass2016controllable}.
MarioGANs~\cite{torrado2019bootstrapping} are specifically designed for level
generation, but they only constrain the mixture of tiles appearing in the
level.  This technique cannot be easily generalized to arbitrary constraints.

In the following, we show how the semantic loss can be used to encode
useful hard constraints in the context of videogame level
generation. These constraints might be functional requirements that
apply to every generated object or might be contextually used to steer
the generation towards objects with certain properties. In our
empirical analysis, we focus on \emph{Super Mario Bros} (SMB),
possibly one of the most studied video games in tile-based level
generation.

Recently, \cite{volz2018evolving} applied Wasserstein GANs
(WGANs)~\cite{arjovsky2017wasserstein} to SMB level generation. The
approach works by first training a generator in the usual
way, then using an evolutionary algorithm called Covariance
Matrix Adaptation Evolution Strategy (CMA-ES) to search for the best
latent vectors according to a user-defined fitness function on the
corresponding levels. We stress that this technique is orthogonal to
CANs and the two can be combined together.
We adopt the same experimental setting, WGAN architecture and training
procedure of~\cite{volz2018evolving}. The structured objects are $14
\times 28$ tile-based representations of SMB levels
(e.g. Fig.~\ref{fig:smb-ex}) and the training data is obtained by
sliding a $28$ tiles window over levels from the \emph{Video game
  level corpus}~\cite{summerville2016vglc}.

We run all the experiments on a machine with a single 1080Ti GPU for $4$ times
with random seeds.


\subsubsection{CANs with tractable constraints: generating SMB levels with \textit{pipes}}
\begin{figure*}[tb]
\begin{center}
\begin{tabular}{cccc}
  {\small \texttt{GAN - pipes}} &
  {\small \texttt{CAN - pipes}} &
  {\small  \texttt{GAN - playable}} &
  {\small \texttt{CAN - playable}} \\
  \includegraphics[width=0.22\linewidth]{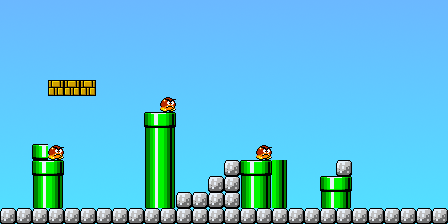} &
  \includegraphics[width=0.22\linewidth]{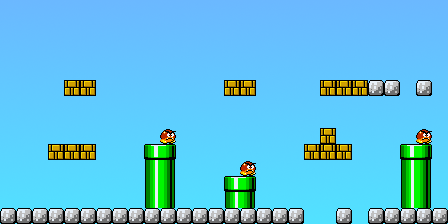} &
  \includegraphics[width=0.22\linewidth]{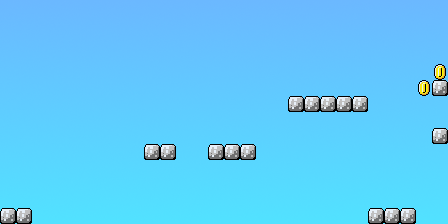} &
  \includegraphics[width=0.22\linewidth]{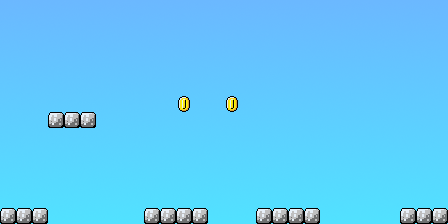} \\
\end{tabular}
\end{center}
\caption{Examples of SMB levels generated by GAN and CAN. Left: generating levels containing pipes; right: generating reachable levels. For each of the two settings we report prototypical examples of levels generated by GAN (first and third picture) and CAN (second and fourth picture). Notice how all pipes generated by CAN are valid, contrarily to what happens for GAN, and that the GAN generates a level that is not playable (because of the big jump at the start of the map).} 

\label{fig:smb-ex}
\end{figure*}

In this experiment, the focus is on showing how CANs can effectively
deal with constraints that can be directly encoded over the generator
output.
Pipes are made of four different types of tiles. They can have a
variable height but the general structure is always the same: two
tiles (\textit{top-left} and \textit{top-right}) on top and one or
more pairs of body tiles (\textit{body-left} and \textit{body-right})
below (see the \texttt{CAN - pipes} in picture in Fig.~\ref{fig:smb-ex} for
examples of valid pipes). Since encoding all possible dispositions and
combinations of pipes in a level would result in an extremely large
propositional formula, we apply the constraint locally to a
$2 \times 2$ window that is slid, horizontally and vertically, by one
tile at a time (notice that all structural properties of pipes are
covered using this method).  The constraint consists of a lot of
implications of the type ``if this is a \textit{top-left} tile, then
the tile below must be a \textit{body-left} one'' conjoined together
(see the Supplementary material for the full formula). The relative
importance of the constraints is determined by the hyper-parameter
$\lambda$ (see Eq.~\ref{eq:can}).

There are two major problems in the application of the constraint on pipes when using a large $\lambda$:
i)~\textit{vanishing pipes}: this occurs because the generator can satisfy the constraint by simply generating layers without pipes;
%
ii)~\emph{mode collapse}: the generator may learn to place pipes always in the same positions.
We address both issues by introducing the SL after an initial
bootstrap phase (of $5,000$ epochs) in which the generator learns to
generate sensible objects, and by linearly increasing its weight from
zero to $\lambda=0.2$. The final value for $\lambda$ was chosen as the
highest value allowing to retain al least 80\% of pipe tiles on
average with respect to a plain GAN. All experiments were run for
$12,000$ epochs.

Table~\ref{tab:pipes-res} reports experimental results comparing GAN
and CAN trained on all levels containing pipes. CAN
manage to almost double the validity of the generated levels (see the
two left pictures in Fig.~\ref{fig:smb-ex} for some prototypical
examples) while retaining about 82\% of the pipe tiles and without any
significant loss in terms of diversity (as measured by the L1 norm on
the difference between each pair of levels in the generated batch) or
cost in terms of training (roughly doubled training times). 
Inference is real-time (< 40 ms) for both architectures.

These results allow to answer {\bf Q1} affirmatively. 

\begin{table*}[tb]
	\centering
	\tiny 
	\begin{tabular}{cccccc}
	\toprule
	\textbf{Model} & \textbf{\# Maps} & \textbf{Validity (\%)} &
	    \textbf{Average pipe-tiles / level} & \textbf{L1 Norm} & \textbf{Training time} \\
    \midrule
	GAN & 7 & 47.6 $\pm$ 8.3 & 7.8   & 0.0115    & 1h 12m \\
	CAN & 7 & 83.2 $\pm$ 4.8 & 6.4   & 0.0110    & 2h 2m \\
	\bottomrule
\end{tabular}

\caption{Comparison between GAN and CAN on SMB level generation with pipes. The 7 maps containing pipes are \textit{mario-1-1}, \textit{mario-2-1}, \textit{mario-3-1},  \textit{mario-4-1}, \textit{mario-4-2}, \textit{mario-6-2} and \textit{mario-8-1}, for a total of $1,404$ training samples. Results report validity, average number of pipe tiles per level, L1 norm on the difference between each pair of levels in the generated batch and training time. Inference is real-time (< 40 ms) for both architectures.}
\label{tab:pipes-res}
\end{table*}

\subsubsection{CANs with intractable constraints: generating \textit{playable} SMB levels}

In the following we show how CANs can be successfully applied in
settings where constraints are too complex to be directly encoded onto
the generator output. A level is \emph{playable} if there is a
feasible path\footnote{According to the game's physics.} from the
left-most to the right-most column of the level. We refer to this
property as {\em reachability}. We compare CANs with CMA-ES, as both
techniques can be used to steer the network towards the generation of
playable levels.
In CMA-ES, the fitness function doesn't have to be differentiable and
the playability is computed on the output of an A* agent (the same used in ~\cite{volz2018evolving}) playing
the level. Having the SL to steer the generation towards playable
levels is not trivial, since it requires a differentiable definition
of playability.
Directly encoding the constraint in propositional logic is
intractable. Consider the size of a first order logic propositional formula
describing all possible path a player can follow in the level. We thus define 
the playability constraint on the output of an embedding function $\phi$
(modelled as a feedforward NN) that approximates tile
reachability. The function is trained to predict whether each tile is
reachable from the left-most column using traces obtained from the A*
agent. See the Supplementary material for the details.

\begin{table*}[tb]
	\centering
	\tiny
	\begin{tabular}{cccccc}
	\toprule
	\textbf{Network type} & \textbf{Level} & \textbf{Tested samples} & \textbf{Validity} & \textbf{Training time} & \textbf{Inference time per sample} \\
	\midrule
	GAN &			mario-1-3	&	1000 &	9.80\% &		1 h 15 min &	$\sim$ 40 ms \\
	GAN + CMA-ES &	mario-1-3	&	1000 &	65.90\% &		1 h 15 min &	$\sim$ 22 min \\
	CAN &			mario-1-3	&	1000 &	71.60\% &	 	1 h 34 min & 	$\sim$ 40 ms \\
	\midrule
	GAN &			mario-3-3	&	1000 &	13.00\% &		1 h 11 min &	$\sim$ 40 ms \\
	GAN + CMA-ES &	mario-3-3	&	1000 &	64.20\% &		1 h 11 min & 	$\sim$ 22 min \\
	CAN &			mario-3-3   &	1000 &	62.30\% &		1 h 27 min & 	$\sim$ 40 ms \\
	\bottomrule
\end{tabular}
\caption{Results on the generation of {\em playable} SMB level. Levels
  \textit{mario-1-3} ($123$ training samples) and \textit{mario-3-3}
  ($122$ training samples) were chosen due to their
  high solving complexity. Results compare a baseline GAN, a GAN
  combined with CMA-ES and a CAN. Validity is defined as the ability
  of the A* agent to complete the level. Note that inference time for
  GAN and CAN is measured in milliseconds while time for GAN + CMA-ES
  is in minutes.}
\label{tab:smb-res}
\end{table*}

Table~\ref{tab:smb-res} shows the validity of a batch of $1,000$
levels generated respectively by plain GAN, GAN combined
with CMA-ES using the default parameters for the search, and a forward
pass of CAN. Each training run lasted
$15000$ epochs with all the default hyper parameters defined
in~\cite{volz2018evolving}, and the SL was activated from epoch $5000$
with $\lambda = 0.01$, which validation experiments showed to be a
reasonable trade-off between SL and generator loss.
Results show that CANs achieves better (\textit{mario-1-3}) or
comparable (\textit{mario-3-3}) validity with respect to GAN +
CMA-ES at a fraction of the inference time. At the cost of pretraining
the reachability function, CANs avoid the execution of the A* agent
during the generation and sample high quality objects in milliseconds
(as compared to minutes), thus enabling applications to create new
levels at run time. Moreover, no significant quality degradation can
be seen on the generated levels as compared to the ones generated by
plain GAN (which on the other hand fails most of the time to
generate reachable levels), as can be seen in
Fig.~\ref{fig:smb-ex}.
With these results, we can answer \textbf{Q2} affirmatively.


\subsection{Molecule generation}
\label{sec:molecules}



Most approaches in molecule generation use variational autoencoders
(VAEs)~\cite{gomez2018automatic,kusner2017grammar,dai2018syntax,samanta2019nevae},
or more expensive techniques like
MCMC~\cite{seff2019discrete}. Closest to CANs are
ORGANs~\cite{guimaraes2017objective} and
MolGANs~\cite{de2018molgan}, which respectively combine Sequence GANs
(SeqGANs) and Graph Convolutional Networks (GCNs) with a reward
network that optimizes specific chemical properties.
Albeit comparing favorably with both sequence
models~\cite{jaques2017sequence,guimaraes2017objective} (using SMILE
representations) and likelihood-based methods, MolGAN are reported to
be susceptible to mode collapse.

%
In this experiment, we investigate \textbf{Q3} by combining MolGAN's
adversarial training and \emph{reinforcement learning objective} with
a conditional SL term on the task of generating molecules with certain
desirable chemical properties.
%
%
%
In contrast with our previous experimental settings, here the
structured objects are undirected graphs of bounded maximum size,
represented by discrete tensors that encode the atom/node type
(padding atom (no atom), Carbon, Nitogren,
Oxygen, Fluorine) and the bound/edge type
(padding bond (no bond), single, double,
triple and aromatic bond).
%
%
%
%
%
%
%
During training, the network implicitly rewards validity and the
maximization of the three chemical properties at once: \textbf{QED}
(druglikeness), \textbf{SA} (synthesizability) and \textbf{logP}
(solubility). The training is stopped once the uniqueness drops under
$0.2$.
We augment the MolGAN architecture with a conditional SL term, making
use of $4$ latent dimensions to control the presence of one of the $4$
types of atoms considered in the experiment, as shown in
Section~\ref{sec:infocan}.

%
Conditioning the generation of molecules with specific atoms at
training time mitigates the drop in uniqueness caused by the reward
network during the training. This allows the model to be trained for
more epochs and results in more diverse and higher quality molecules, as reported in
Table~\ref{tab:mol}.

%
%

In this experiment, we train the model on a NVIDIA RTX 2080 Ti. The total
training time is around 1 hour, and the inference is real-time. Using CANs produced a negligible overhead during the training with respect to the original model, providing further evidence that the technique doesn't heavily impact on the training.
This results suggest that coupling CANs with a reinforcement learning
objective is beneficial, answering \textbf{Q3} affirmatively.

\begin{table*}[tb]
    \centering
    \begin{footnotesize}
    \begin{tabular}{cccccccc}
    \toprule
        \textbf{Reward for} & \textbf{SL} & \textbf{validity} & \textbf{uniqueness} & \textbf{diversity} & \textbf{QED} & \textbf{SA}   & \textbf{logP} \\ \midrule
        \multirow{2}*{QED + SA + logP}
                    & False               & 97.4              & 2.4                 & 91.0               & 47.0         & 84.0          & 65.0          \\ \cline{2-8}
                    & True                & 96.6      & 2.5         & 98.8       & 51.8 & 90.7  & 73.6  \\ \bottomrule
    \end{tabular}
    \end{footnotesize}
\caption{Results of using the semantic loss on the MolGAN
  architecture. The diversity score is obtained by comparing
  sub-structures of generated samples against a random subset of the
  dataset. A lower score indicates a higher amount of repetitions
  between the generated samples and the dataset.  The first row refers
  to the results reported in the MolGAN paper.
}
\label{tab:mol}
\end{table*}


\section{Conclusion}

We presented Constrained Adversarial Networks (CANs), a generalization of GANs in
which the generator is encouraged \emph{during training} to output valid
structures.  CANs make use of the semantic loss~\cite{xu2018semantic} to
penalize the generator proportionally to the mass it allocate to invalid
structures and.  As in GANs, generating valid structures (on average)
requires a simple forward pass on the generator.  Importantly, the data
structures used by the SL, which can be large if the structural constraints
are very complex, are discarded after training.
CANs were proven to be effective in improving the quality of the
generated structures without significantly affecting inference
run-time, and conditional CANs proved useful in promoting diversity of
the generator's outputs.
%
%


\section*{Broader Impact}

Broadly speaking, this work aims at improving the reliability of
structures / configurations generated via machine learning
approaches. This can have a strong impact on a wide range of research
fields and application domains, from drug design and protein
engineering to layout synthesis and urban planning. Indeed, the lack
of reliability of machine-generated outcomes is one of main obstacles
to a wider adoption of machine learning technology in our
societies. On the other hand, there is a risk of overestimating the
reliability of the outputs of CANs, which are only guaranteed to
satisfy constraints in expectation. 
For applications in which invalid structures should be avoided, like
safety-critical applications, the objects output by CANs should always
be validated before use.

From an artificial intelligence perspective, this work supports the
line of thought that in order to overcome the current limitations of
AI there is a need for combining machine learning and especially deep
learning technology with approaches from knowledge representation and
automated reasoning, and that principled ways to achieve this
integration should be pursued.

\begin{ack}
    We are grateful to Gianvito Taneburgo for carrying out some preliminary experiments and to the anonymous reviewers for helping to improve the manuscript.  This work has received funding from the European Research Council (ERC) under the European Union’s Horizon 2020 research and innovation programme (grant agreement No.  [694980] SYNTH: Synthesising Inductive Data Models).  The research of ST has received funding from the ``DELPhi - DiscovEring Life Patterns'' project funded by the MIUR Progetti di Ricerca di Rilevante Interesse Nazionale (PRIN) 2017 -- DD n. 1062 del 31.05.2019. The research of AP was partially supported by TAILOR, a project funded by EU Horizon 2020 research and innovation programme under GA No 952215.
\end{ack}



\bibliographystyle{unsrt}
\bibliography{paper}
\end{document}